\documentclass{article}

\PassOptionsToPackage{numbers, compress}{natbib}
\usepackage{microtype}
\usepackage{graphicx}
\usepackage{booktabs} 

\usepackage{xr} 
\usepackage{hyperref}


\usepackage[preprint]{neurips_2019}

\usepackage[utf8]{inputenc} 
\usepackage[T1]{fontenc}    
\usepackage{url}            
\usepackage{amsfonts}       
\usepackage{nicefrac}       
\usepackage{xfrac}       

\usepackage[dvipsnames,table]{xcolor}
\usepackage{pgfplotstable}
\usepackage{tikz}
\usepackage{pgfplots}
\usepackage{collcell}
\usepackage{xstring}
\usepackage{tabularx}
\usepackage{tabu}
\usepackage{environ}
\usepackage{fp}
\usepackage{upgreek}
\usepackage{amsmath}
\usepackage{amsthm}
\usepackage{algorithm, algorithmic}
\usepackage{enumitem}
\usepackage{bm}
\usepackage{subfiles}
\usepackage{multirow}
\usepackage{float}
\usepackage{dblfloatfix}
\usepackage{placeins}
\usepackage{stackengine}
\usepackage{caption,subcaption}



\newcommand{\btheta}{\bm{\theta}}
\newcommand{\bX}{\bm{X}}
\newcommand{\bY}{\bm{Y}}
\newcommand{\Ncol}{N_\mathrm{col}}
\newcommand{\Nrow}{N_\mathrm{row}}


\newcommand{\TODO}[1]{}
\newcommand{\kevin}[1]{}
\newcommand{\yarin}[1]{}
\newcommand{\geoff}[1]{}

\NewEnviron{scaleblock65}{\resizebox{0.65\columnwidth}{!}{
\vbox{\BODY}
}}

\NewEnviron{scaleblock75}{\resizebox{0.75\columnwidth}{!}{
\vbox{\BODY}
}}

\pgfplotstableset{
    color cells resize75/.style={
        font={\small},
        format=inline,
        begin table=\begin{scaleblock75}\begin{tabular},
        end table=\end{tabular}\end{scaleblock75},
        col sep=comma,
        string type,
        postproc cell content/.code={%
                \pgfkeysalso{@cell content=\ifthenelse{\equal{##1}{n}}{\,}{\rule{0cm}{2.4ex}\cellcolor{black!##1}\pgfmathtruncatemacro\number{##1}\ifnum\number>40\color{white}\fi##1}}%
                },
        columns/x/.style={
            column name={},
            postproc cell content/.code={}
            },
        columns/n/.style={
            column name={},
            },
        columns/o/.style={
            column name={},
            },
        columns/m/.style={
            column name={},
            postproc cell content/.code={##1}
            },
        columns/m2/.style={
            column name={},
            postproc cell content/.code={##1}
            },
        columns/none2/.style={
            column name={none}
            },
        columns/t5050/.style={
            column name={$\underset{\alpha=0.5, \gamma=0.5}{\text{targeted}}$},
            },
        columns/t50502/.style={
            column name={$\underset{\alpha=0.5, \gamma=0.5}{\text{targeted}}$},
            },
        columns/tu5050/.style={
            column name={$\underset{\alpha=0.5, \gamma=0.5}{\text{targeted unit}}$},
            },
        columns/tw5050/.style={
            column name={$\underset{\alpha=0.5, \gamma=0.5}{\text{targeted weight}}$},
            },
        columns/t5050l1/.style={
            column name={$\underset{\alpha=0.5, \gamma=0.5}{\text{targeted}}+L^1_{0.1}$},
            },
        columns/t7533/.style={
            column name={$\underset{\alpha=0.33, \gamma=0.75}{\text{targeted}}$},
            },
        columns/t7566/.style={
            column name={$\underset{\alpha=0.66, \gamma=0.75}{\text{targeted}}$},
            },
        columns/t7590/.style={
            column name={$\underset{\alpha=0.90, \gamma=0.75}{\text{targeted}}$},
            },
        columns/t9066/.style={
            column name={$\underset{\alpha=0.66, \gamma=0.90}{\text{targeted}}$},
            },
        columns/t9075/.style={
            column name={$\underset{\alpha=0.75, \gamma=0.90}{\text{targeted}}$},
            },
        columns/t9088/.style={
            column name={$\underset{\alpha=0.88, \gamma=0.90}{\text{targeted}}$},
            },
        columns/w25/.style={
            column name={$\underset{\alpha=0.25}{\text{dropout}}$},
            },
        columns/u25/.style={
            column name={$\underset{\alpha=0.25}{\text{dropout}}$},
            },
        columns/var_weight/.style={
            column name={variational},
            },
        columns/var_unit/.style={
            column name={variational},
            },
        columns/smallify1en4/.style={
            column name={$\underset{\lambda=0.0001}{\text{smallify}}$},
            },
        columns/smallify1en51/.style={
            column name={$\underset{\lambda=0.00001}{\text{smallify}}$},
            },
        columns/smallify1en52/.style={
            column name={$\underset{\lambda=0.00001}{\text{smallify}}$},
            },
        columns/ramptarg1/.style={
            column name={$\underset{\alpha=0.99, \gamma=0.99}{\text{ramp targ}}$},
            },
        columns/ramptarg2/.style={
            column name={$\underset{\alpha=0.99, \gamma=0.99}{\text{ramp targ}}$},
            },
        columns/ramptarg3/.style={
            column name={$\underset{\alpha=0.99, \gamma=0.90}{\text{ramp targ}}$},
            },
        every column/.style={
            column name={##1}
        }
    },
        color cells resize65/.style={
        font={\small},
        format=inline,
        begin table=\begin{scaleblock65}\begin{tabular},
        end table=\end{tabular}\end{scaleblock65},
        col sep=comma,
        string type,
        postproc cell content/.code={%
                \pgfkeysalso{@cell content=\ifthenelse{\equal{##1}{n}}{\,}{\rule{0cm}{2.4ex}\cellcolor{black!##1}\pgfmathtruncatemacro\number{##1}\ifnum\number>40\color{white}\fi##1}}%
                },
        columns/x/.style={
            column name={},
            postproc cell content/.code={}
            },
        columns/n/.style={
            column name={},
            },
        columns/o/.style={
            column name={},
            },
        columns/m/.style={
            column name={},
            postproc cell content/.code={##1}
            },
        columns/m2/.style={
            column name={},
            postproc cell content/.code={##1}
            },
        columns/t5050/.style={
            column name={$\underset{\alpha=0.5, \gamma=0.5}{\text{targeted}}$},
            },
        columns/tu5050/.style={
            column name={$\underset{\alpha=0.5, \gamma=0.5}{\text{targeted unit}}$},
            },
        columns/tw5050/.style={
            column name={$\underset{\alpha=0.5, \gamma=0.5}{\text{targeted weight}}$},
            },
        columns/t5050l1/.style={
            column name={$\underset{\alpha=0.5, \gamma=0.5}{\text{targeted}}+L^1_{0.1}$},
            },
        columns/t7533/.style={
            column name={$\underset{\alpha=0.33, \gamma=0.75}{\text{targeted}}$},
            },
        columns/t7566/.style={
            column name={$\underset{\alpha=0.66, \gamma=0.75}{\text{targeted}}$},
            },
        columns/t7590/.style={
            column name={$\underset{\alpha=0.90, \gamma=0.75}{\text{targeted}}$},
            },
        columns/t9075/.style={
            column name={$\underset{\alpha=0.75, \gamma=0.90}{\text{targeted}}$},
            },
        columns/t9022/.style={
            column name={$\underset{\alpha=0.22, \gamma=0.90}{\text{targeted}}$},
            },
        columns/w25/.style={
            column name={$\underset{\alpha=0.25}{\text{dropout}}$},
            },
        columns/u25/.style={
            column name={$\underset{\alpha=0.25}{\text{dropout}}$},
            },
        columns/var_weight/.style={
            column name={variational},
            },
        columns/var_unit/.style={
            column name={variational},
            },
        columns/smallify/.style={
            column name={$\underset{\lambda=0.0001}{\text{smallify}}$},
            },
        every column/.style={
            column name={##1}
        }
    },
    color cells/.style={
        font={\small},
        format=inline,
        begin table=\begin{tabular},
        end table=\end{tabular},
        col sep=comma,
        string type,
        postproc cell content/.code={%
                \pgfkeysalso{@cell content=\ifthenelse{\equal{##1}{n}}{\,}{\rule{0cm}{2.4ex}\cellcolor{black!##1}\pgfmathtruncatemacro\number{##1}\ifnum\number>40\color{white}\fi##1}}%
                },
        columns/x/.style={
            column name={},
            postproc cell content/.code={}
            },
        columns/n/.style={
            column name={},
            },
        columns/o/.style={
            column name={},
            },
        columns/m/.style={
            column name={},
            postproc cell content/.code={##1}
            },
        columns/m2/.style={
            column name={},
            postproc cell content/.code={##1}
            },
        columns/t5050/.style={
            column name={$\underset{\alpha=0.5, \gamma=0.5}{\text{targeted}}$},
            },
        columns/tu5050/.style={
            column name={$\underset{\alpha=0.5, \gamma=0.5}{\text{targeted unit}}$},
            },
        columns/tw5050/.style={
            column name={$\underset{\alpha=0.5, \gamma=0.5}{\text{targeted weight}}$},
            },
        columns/w25/.style={
            column name={$\underset{\alpha=0.25}{\text{weight}}$},
            },
        columns/u25/.style={
            column name={$\underset{\alpha=0.25}{\text{unit}}$},
            },
        columns/var_weight/.style={
            column name={variational},
            },
        columns/var_unit/.style={
            column name={variational},
            },
        columns/smallify/.style={
            column name={$\underset{\lambda=0.0001}{\text{smallify}}$},
            },
        columns/t7533/.style={
            column name={$\underset{\alpha=0.33, \gamma=0.75}{\text{targeted}}$},
            },
        columns/t7566/.style={
            column name={$\underset{\alpha=0.66, \gamma=0.75}{\text{targeted}}$},
            },
        columns/t9075/.style={
            column name={$\underset{\alpha=0.75, \gamma=0.90}{\text{targeted}}$},
            },
        columns/t9022/.style={
            column name={$\underset{\alpha=0.22, \gamma=0.90}{\text{targeted}}$},
            },
        every column/.style={
            column name={##1}
        }
    }
}

\title{Learning Sparse Networks Using Targeted Dropout}
\author{%
    Aidan N. Gomez\,\(^{1,2,3}\) \qquad Ivan Zhang\,\(^{2}\)
    \AND
    Siddhartha Rao Kamalakara\,\(^{2}\)  \qquad Divyam Madaan\,\(^{2}\)
    \AND
    Kevin Swersky\,\(^{1}\) \qquad Yarin Gal\,\(^{3}\) \qquad Geoffrey E. Hinton\,\(^{1}\) \\
    \AND
    \\
    \(^1\)\,Google Brain
    \And
    \\
    \(^2\)\,for.ai
    \And
    \\
    \(^3\)\,Department of Computer Science \\
    University of Oxford
}
\begin{document}

\maketitle

\vspace{-3mm}
\begin{abstract}
Neural networks are easier to optimise when they have many more weights than are required for modelling the mapping from inputs to outputs. This suggests a two-stage learning procedure that first learns a large net and then prunes away connections or hidden units. But standard training does not necessarily encourage nets to be amenable to pruning. We introduce targeted dropout, a method for training a neural network so that it is robust to subsequent pruning. Before computing the gradients for each weight update, targeted dropout stochastically selects a set of units or weights to be dropped using a simple self-reinforcing sparsity criterion and then computes the gradients for the remaining weights.  The resulting network is robust to {\it post hoc} pruning of weights or units that frequently occur in the dropped sets. The method improves upon more complicated sparsifying regularisers while being simple to implement and easy to tune.
\end{abstract}

\section{Introduction}
Neural networks are a powerful class of models that achieve the state-of-the-art on a wide range of tasks such as object recognition, speech recognition, and machine translation. One reason for their success is that they are extremely flexible models because they have a large number of learnable parameters. However, this flexibility can lead to overfitting, and can unnecessarily increase the computational and storage requirements of the network.

There has been a large amount of work on developing strategies to compress neural networks. One intuitive strategy is \emph{sparsification}: removing weights or entire units from the network. Sparsity can be encouraged during learning by the use of sparsity-inducing regularisers, like $L^1$ or $L^0$ penalties. It can also be imposed by \emph{post hoc} pruning, where a full-sized network is trained, and then sparsified according to some pruning strategy. Ideally, given some measurement of task performance, we would prune the weights or units that provide the least amount of benefit to the task. Finding the optimal set is, in general, a difficult combinatorial problem, and even a greedy strategy would require an unrealistic number of task evaluations, as there are often millions of parameters. Common pruning strategies therefore focus on fast approximations, such as removing weights with the smallest magnitude \cite{han2015learning}, or ranking the weights by the sensitivity of the task performance with respect to the weights, and then removing the least-sensitive ones \cite{lecun1990optimal}. The hope is that these approximations correlate well with task performance, so that pruning results in a highly compressed network while causing little negative impact to task performance, however this may not always be the case.

Our approach is based on the observation that dropout regularisation \citep{hinton2012improving,srivastava2014dropout} itself enforces sparsity tolerance during training, by sparsifying the network with each forward pass. This encourages the network to learn a representation that is robust to a particular form of post hoc sparsification -- in this case, where a random set of units is removed. Our hypothesis is that if we plan to do explicit post hoc sparsification, then we can do better by specifically applying dropout to the set of units that we a priori believe are the least useful. We call this approach \emph{targeted dropout}. The idea is to rank weights or units according to some fast, approximate measure of importance (like magnitude), and then apply dropout primarily to those elements deemed unimportant. Similar to the observation with regular dropout, we show that this encourages the network to learn a representation where the importance of weights or units more closely aligns with our approximation. In other words, the network learns to be robust to \emph{our choice} of post hoc pruning strategy.

The advantage of targeted dropout as compared to other approaches is that it makes networks extremely robust to the {\it post hoc} pruning strategy of choice, gives intimate control over the desired sparsity patterns, and is easy to implement~\footnote{Code available at: \url{github.com/for-ai/TD}
, as well as in 
Tensor2Tensor \citep{vaswani2018tensor2tensor}
}, consisting of a two-line change for neural network frameworks such as Tensorflow \citep{tensorflow} or PyTorch \citep{pytorch}. The method achieves impressive sparsity rates on a wide range of architectures and datasets; notably 99\% sparsity on the ResNet-32 architecture for a less than 4\% drop in test set accuracy on CIFAR-10.

\section{Background}

In order to present targeted dropout, we first briefly introduce some notation, and review the concepts of dropout and magnitude-based pruning.

\newcommand{\weights}{\Omega_{\bm{\theta}}}

\subsection{Notation}
Assume we are dealing with a particular network architecture. We will use $\bm{\theta} \in \Theta$ to denote the vector of parameters of a neural network drawn from candidate set \(\Theta\), with $|\bm{\theta}|$ giving the number of parameters. \(\weights\) denotes the set of weight matrices in a neural network parameterised by \(\bm{\theta}\), accordingly, we will denote $\mathbf{W} \in \weights$ as a weight matrix that connects one layer to another in the network. We will only consider weights, ignoring biases for convenience, and note that biases are not removed during pruning. For brevity, we will use the notation $\bm{w}_o \equiv \mathbf{W}_{\cdot, o}$ to denote the weights connecting the layer below to the $o^\mathrm{th}$ output unit (i.e. the $o^\mathrm{th}$ column of the weight matrix), $\Ncol(\mathbf{W})$ to denote the number of columns in $\mathbf{W}$, and $\Nrow(\mathbf{W})$ to denote the number of rows. Each column corresponds to a hidden unit, or feature map in the case of convolutional layers. Note that flattening and concatenating all of the weight matrices in \(\weights\) would recover \(\bm{\theta}\).

\subsection{Dropout}\label{sec:dropout}
Our work uses the two most popular Bernoulli dropout techniques, \citeauthor{hinton2012improving}'s unit dropout \citep{hinton2012improving,srivastava2014dropout} and \citeauthor{wan2013regularization}'s weight dropout (dropconnect) \citep{wan2013regularization}.
For a fully-connected layer with input tensor \(\bX\), weight matrix \(\mathbf{W}\), output tensor \(\bY\), and mask \(\mathbf{M} \sim \operatorname{Bernoulli}(1-\alpha)\) we define both techniques below:

\textbf{Unit dropout} \citep{hinton2012improving,srivastava2014dropout}:
\[\bY=(\bX \odot \mathbf{M})\mathbf{W}\]
Unit dropout randomly drops \textit{units} (often referred to as neurons) at each training step to reduce dependence between units and prevent overfitting.

\textbf{Weight dropout} \citep{wan2013regularization}:
\[\bY = \bX(\mathbf{W} \odot \mathbf{M})\]
Weight dropout randomly drops individual \textit{weights} in the weight matrices at each training step. Intuitively, this is dropping connections between layers, forcing the network to adapt to a different connectivity at each training step.

\subsection{Magnitude-based pruning}\label{sec:pruning}

A popular class of pruning strategies are those characterised as \textit{magnitude-based} pruning strategies. These strategies treat the top-\(k\) largest magnitude weights as important. We use $\operatorname{argmax-k}$ to return the top-\(k\) elements (units or weights) out of all elements being considered.

\textbf{Unit pruning} \citep{molchanov2016pruning,frankle2018lottery}: considers the units (column-vectors) of weight matrices under the \(L^2\)-norm.

\begin{equation}\label{eqn:unitpruning}
\mathcal{W}(\btheta)=\left\{\underset{\substack{\bm{w}_o \\ 1 \leq o \leq \Ncol(\mathbf{W})}}{\operatorname{argmax-k}} \,\|\bm{w}_o\|_2 \,\bigg|\, \mathbf{W}\in\weights \right\}
\end{equation}

\textbf{Weight pruning} \citep{han2015learning,molchanov2016pruning}: considers the entries of each feature vector under the \(L^1\)-norm. Note that the top-$k$ is with respect to the other weights within the same feature vector.

\begin{equation}\label{eqn:weightpruning}
\mathcal{W}(\btheta)=\left\{\underset{\substack{\mathbf{W}_{io} \\ 1 \leq i \leq \Nrow(\mathbf{W})}}{\operatorname{argmax-k}} \,|\mathbf{W}_{io}| \,\bigg|\, 1 \leq o \leq \Ncol(\mathbf{W}), \mathbf{W}\in\weights \right\}
\end{equation}

While weight pruning tends to preserve more of the task performance under coarser prunings \citep{han2015deep,ullrich2017soft,frankle2018lottery}, unit pruning allows for considerably greater computational savings \citep{wen2016learning,louizos2017learning}. In particular, weight pruned networks can be implemented using sparse linear algebra operations, which offer speedups only under sufficiently sparse conditions; while unit pruned networks execute standard linear algebra ops on lower dimensional tensors, which tends to be a much faster option for given a fixed sparsity rate.

\section{Targeted Dropout}

Consider a neural network parameterized by \(\bm{\theta}\), and our importance criterion (defined above in Equations~\eqref{eqn:unitpruning} and \eqref{eqn:weightpruning}) \(\mathcal{W}(\btheta)\). We hope to find optimal parameters \(\btheta^*\) such that our loss \(\mathcal{E}(\mathcal{W}(\btheta^*))\) is low, and at the same time \(|\mathcal{W}(\btheta^*)| \leq k \), i.e.\ we wish to keep only the $k$ weights of highest magnitude in the network. A deterministic pruning implementation would select the bottom $|\btheta| - k$ elements and drop them out. However, we would like for low-valued elements to be able to increase their value if they become important during training. Therefore, we introduce stochasticity into the process using a targeting proportion $\gamma$ and a drop probability $\alpha$. The targeting proportion means that we select the bottom $\gamma|\btheta|$ weights as candidates for dropout, and of those we drop the elements independently with drop rate $\alpha$. This implies that the expected number of units to keep during each round of targeted dropout is $(1 - \gamma \cdot \alpha) |\btheta|$. As we will see below, the result is a reduction in the important subnetwork's dependency on the unimportant subnetwork, thereby reducing the performance degradation as a result of pruning at the conclusion of training.

\subsection{Dependence Between the Important and Unimportant Subnetworks}

The goal of targeted dropout is to reduce the dependence of the important subnetwork on its complement. A commonly used intuition behind dropout is the prevention of coadaptation between units; that is, when dropout is applied to a unit, the remaining network can no longer depend on that unit's contribution to the function and must learn to propagate that unit's information through a more reliable channel. An alternative description asserts that dropout maximizes the mutual information between units in the same layer, thereby decreasing the impact of losing a unit~\cite{srivastava2014dropout}. Similar to our approach, dropout can be used to guide properties of the representation. For example, nested dropout~\citep{rippel2014learning} has been shown to impose `hierarchy' among units depending on the particular drop rate associated with each unit. Dropout itself can also be interpreted as a Bayesian approximation \citep{Gal2016Uncertainty}.

A more relevant intuition into the effect of targeted dropout in our specific pruning scenario can be obtained from an illustrative case where the important subnetwork is completely separated from the unimportant one. Suppose a network was composed of two non-overlapping subnetworks, each able to produce the correct output by itself, with the network output given as the average of both subnetwork outputs. If our importance criterion designated the first subnetwork as important, and the second subnetwork as unimportant (more specifically, it has lower weight magnitude), then adding noise to the weights of the unimportant subnetwork (i.e.\ applying dropout) means that with non-zero probability we will corrupt the network output. Since the important subnetwork is already able to predict the output correctly, to reduce the loss we must therefore reduce the weight magnitude of the unimportant subnetwork output layer towards zero, in effect ``killing'' that subnetwork, and reinforcing the separation between the important subnetwork and the unimportant one.

These interpretations make clear why dropout should be considered a natural tool for application in pruning.
We can empirically confirm targeted dropout's effect on weight dependence by comparing a network trained with and without targeted dropout and inspecting the Hessian and gradient to determine the dependence of the network on the weights/units to be pruned. As in \citet{lecun1990optimal}, we can estimate the effect of pruning weights by considering the second degree Taylor expansion of change in loss, $\Delta\mathcal{E} = |\mathcal{E}(\btheta - \bm{d})-\mathcal{E}(\btheta)|$:
\begin{align}\label{eqn:cie}
    \Delta\mathcal{E} &= |-\nabla_{\btheta} \mathcal{E}^\top\bm{d} + \sfrac{1}{2}\,\bm{d}^\top\bm{H}\bm{d} + \mathcal{O}(\|\bm{d}\|^3)|
\end{align}
Where \(d_i = \theta_i\) if \(\theta_i \in \overline{\mathcal{W}(\btheta)}\) (the weights to be removed) and $0$ otherwise.
\(\nabla_{\btheta}\mathcal{E}\) are the gradients of the loss, and \(\bm{H}\) is the Hessian. Note that at the end of training, if we have found a critical point \(\btheta^*\), then \(\nabla_{\btheta}\mathcal{E}(\btheta^*)=\bm{0}\), leaving only the Hessian term. In our experiments we empirically confirm that targeted dropout reduces the dependence between the important and unimportant subnetworks by an order of magnitude (See Fig.\ \ref{fig:hess}, and Section~\ref{sec:loseticket} for more details).

\section{Related Work}

The pruning and sparsification of neural networks has been studied for nearly three decades and has seen a substantial increase in interest due to their implementation on resource limited devices such as mobile phones and ASICs. Early work such as optimal brain damage \citep{lecun1990optimal} and optimal brain surgeon \citep{hassibi1993second}, as well as more recent efforts \citep{molchanov2016pruning,theis2018faster}, use a second order Taylor expansion of the loss around the weights trained to a local minimum to glean strategies for selecting the order in which to prune parameters. \citet{han2015deep} combine weight quantisation with pruning and achieve impressive network compression results, reducing the spatial cost of networks drastically. \citet{dong2017learning} improve the efficiency of the optimal brain surgeon procedure by making an independence assumption between layers. \citet{wen2016learning} propose using Group Lasso \citep{yuan2006model} on convolutional filters and are able to remove up to 6 layers from a ResNet-20 network for a 1\% increase in error.

A great deal of effort has been put towards developing improved pruning heuristics and sparsifying regularizers \citep{lecun1990optimal,hassibi1993second,han2015deep,babaeizadeh2016noiseout,molchanov2016pruning,dong2017learning,louizos2017learning,huang2018learning,theis2018faster}. These are generally comprised of two components: the first is a regularisation scheme incorporated into training to make the important subnetworks easily identifiable to a post hoc pruning strategy; the second is a particular post hoc pruning strategy which operates on a pre-trained network and strips away the unimportant subnetwork. 

The two works most relevant to our own are \(L^0\) regularisation \citep{louizos2017learning} and variational dropout \citep{molchanov2017variational}. \citet{louizos2017learning} use an adaptation of concrete dropout \citep{gal2017concrete} 
on the weights of a network and regularise the drop rates in order to sparsify the network. Similarly, \citet{molchanov2017variational} apply variational dropout \citep{kingma2015variational} to the weights of a network and note that the prior implicitly sparsifies the parameters by preferring large drop rates. 
In addition to our methods being more effective at shrinking the size of the important subnetwork, targeted dropout uses two intuitive hyperparameters, the targeting proportion $\gamma$ and the drop rate $\alpha$, and directly controls sparsity throughout training (i.e., attains a predetermined sparsity threshold). In comparison, \citet{louizos2017learning} uses the Hard-Concrete distribution which adds three hyperparameters and doubles the number of trainable parameters by introducing a unique gating parameter for each model parameter, which determines the Concrete dropout rate;
while \citet{molchanov2016pruning} adds two hyperparameters and doubles the number of trainable parameters. In our experiments we also compare against \(L^1\) regularization \citep{han2015learning} which is intended to drive unimportant weights towards zero. 

Another dropout-based pruning mechanism is that of \citet{wang2017structured}, where a procedure is used to adapt dropout rates towards zero and one (similar to \citet{louizos2017learning} and \citep{molchanov2017variational}). We recommend \citet{galestateofsparsity}'s rigorous analysis of recently proposed pruning procedures for a complete picture of the efficacy of recent neural network pruning algorithms; in particular, it challenges some of the recent claims suggesting pruning algorithms perform about as well as random pruning procedures \citep{crowley2018pruning,liu2018rethinking}.

Targeted dropout itself is reminiscent of nested dropout \citep{rippel2014learning} which applies a structured form of dropout: a chain structure is imposed on units, and children are deterministically dropped whenever their parent is dropped. In effect, each child unit gets a progressively higher marginal drop rate, imposing a hierarchy across the units; similar to both meProp \citep{meprop} and excitation dropout \citep{excitationdropout}. \citet{rippel2014learning} demonstrate the effect using an autoencoder where nested dropout is applied to the code; the result is a model where one can trade off reconstruction accuracy with compute by dropping lower priority elements of the code. Standout \citep{ba2013adaptive} is another similar variant of dropout; in standout, the activation value of a unit determines the drop rate, where high activations values lead to a higher keep probability and vice versa.

The Lottery Ticket Hypothesis of \citet{frankle2018lottery} demonstrates the existence of a subnetwork that -- in isolation, with the rest of the network pruned away -- both dictates the function found by gradient descent, and can be trained to the same level of task performance with, or without, the remaining network. In our notation, a prediction of this ``winning lottery ticket'' is \(\mathcal{W}(\btheta)\); and the effectiveness of our method suggests that one can reduce the size of the winning lottery ticket by regularising the network.

\section{Experiments}

 Our experiments were performed using the original ResNet \citep{resnet}, Wide ResNet \citep{zagoruyko2016wide}, and Transformer \citep{transformer} architectures; applied to the CIAFR-10 \citep{cifar10}, ImageNet \citep{imagenet}, and WMT English-German Translation datasets. For each baseline experiment we verify our networks reach the reported accuracy on the appropriate test set; we report the test accuracy at differing prune percentages and compare different regularisation strategies. In addition, we compare our targeted dropout to standard dropout where the expected number of dropped weights is matched between the two techniques (i.e. the drop rate of standard dropout runs is set to \(\gamma\cdot\alpha\), the proportion of weights to target times the dropout rate). 

 For our pruning procedure, we perform the greedy layer-wise magnitude-base pruning described in Section~\ref{sec:pruning} to all weight matrices except those leading to the logits. In our experiments we compare targeted dropout against the following competitive schemes:
 
 \textbf{\(L^1\) Regularization} \citep{han2015learning}: Complexity cost \(\mathcal{\btheta} = \|\btheta\|_1\) is added to the cost function. The hope being that this term would drive unimportant weights to zero. In our table we denote this loss by \(L^1_\beta\) where \(\beta\) is the cost-balancing coefficient applied to the complexity term. 

 \textbf{\(L^0\) Regularization} \citep{louizos2017learning}: \citeauthor{louizos2017learning} apply an augmentation of Concrete Dropout \citep{gal2017concrete}, called Hard-Concrete Dropout, to the parameters of a neural network. The mask applied to the weights follows a Hard-Concrete distribution where each weight is associated with a gating parameter that determines the drop rate. The use of the Concrete distribution allows for a differentiable approximation to the \(L^0\) cost, so we may directly minimise it alongside our task objective. When sparisfying these networks to a desired sparsity rate, we prune according to the learned keep probabilities (\(\sigma(\log(\bm{\alpha}))\) from \citep{louizos2017learning}), dropping those weights with lowest keep probabilities first.
 
 \textbf{Variational Dropout} \citep{kingma2015variational,molchanov2017variational}: Similar to the technique used for \(L^0\) regularisation, \citet{molchanov2017variational} apply Gaussian dropout with trainable drop rates to the weights of the network and interprets the model as a variational posterior with a particular prior. The authors note that the variational lower bound used in training favors higher drop probabilities and experimentally confirm that networks trained in this way do indeed sparsify.

 \textbf{Smallify} \citep{leclerc2018smallify}: \citeauthor{leclerc2018smallify} use trainable gates on weights/units and regularise gates towards zero using \(L^1\) regularisation. Crucial to the technique is the online pruning condition: Smallify keeps a moving variance of the sign of the gates, and a weight/unit's associated gate is set to zero (effectively pruning that weight/unit) when this variance exceeds a certain threshold. This technique has been shown to be extremely effective at reaching high prune rates on VGG networks \citep{simonyan2014very}.
  
Specifically, we compare the following techniques:
\begin{itemize}[labelwidth=5.25em, leftmargin=5.25em, itemindent=0pt]
    \item[\(\underset{\alpha}{\text{dropout}}\):] Standard weight or unit dropout applied at a rate of \(\alpha\).
    \item[\(\underset{\alpha,\gamma}{\text{targeted}}\):] Targeted dropout (the weight variant in `a)' tables, and unit variant in `b)' tables) applied to the \(\gamma\cdot100\%\) lowest magnitude weights at a rate of \(\alpha\). 
    \item[variational:] Variational dropout \citep{kingma2015variational,molchanov2017variational} applied with a cost coefficient of  \(0.01 / 50,000\).
    \item[\(L^0_{\beta}\):] \(L^0\) regularisation \citep{louizos2017learning} applied with a cost coefficient of \(\beta / 50,000\).
    \item[\(L^1_{\beta}\):] \(L^1\) regularisation \citep{han2015learning} applied with a cost coefficient of \(\beta\).
    \item[\(\underset{\lambda}{\text{smallify}}\):] Smallify SwitchLayers \citep{leclerc2018smallify} applied with a cost coefficient of \(\lambda\), exponential moving average decay of 0.9, and a variance threshold of 0.5.
\end{itemize}

\subsection{Analysing the Important Subnetwork}\label{sec:loseticket}

\begin{figure}[t]
\begin{center}
\includegraphics[width=0.45\columnwidth]{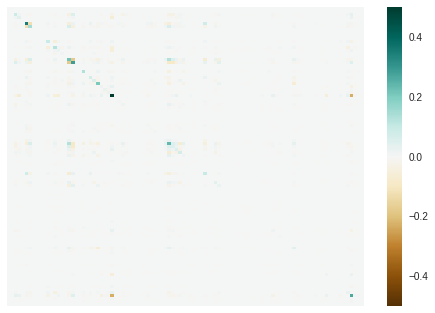}
\hspace{1.2em}\includegraphics[width=0.45\columnwidth]{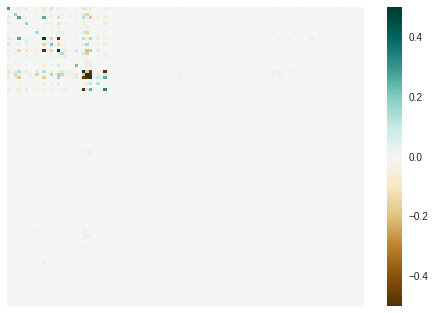}
\end{center}
\begin{center}
\caption{A comparison between a network without dropout (left) and with targeted dropout (right) of the matrix formed by \(\theta^\top\odot\bm{H}\odot\theta\). The weights are ordered such that the last 75\% are the weights with the lowest magnitude (those we intend to prune). 
The sum of the elements of the lower right hand corner approximates the change in error after pruning (Eqn.~\eqref{eqn:cie}).
Note the stark difference between the two networks, with targeted dropout concentrating its dependence on the top left corner, leading to a much smaller error change after pruning (given in Table \ref{tab:alt}).
}
\label{fig:hess}
\end{center}
\vskip -0.2in
\end{figure}

\begin{table}[t]
\begin{center}
\begin{sc}
\begin{tabular}{r|c|c|c}
    \toprule
    Regularisation & \(|\Delta\mathcal{E}|\) & Unpruned Accuracy & Pruned Accuracy \\
    \midrule
    None & 0.120698 & 38.11\% & 26.13\% \\
    Targeted Dropout & 0.0145907 & 40.09\% & 40.14\% \\
    \bottomrule
\end{tabular}
\end{sc}
\end{center}
\caption{Comparison of the change in loss (\(|\Delta\mathcal{E}|\) of Equation~\eqref{eqn:cie}) for dense networks.}\label{tab:alt}
\end{table}

\begin{table*}[t]
\begin{center}
\textbf{Weight Dropout/Pruning} \\
\end{center}
\hspace{3em}
\rotatebox[origin=c]{90}{\small prune percentage}\pgfplotstabletypeset[color cells resize65]{
        x,none,w25,t5050,t7533,t7566,t9075,n,var_weight,$L^1_{0.1}$,$L^0_{0.1}$
        0 \%,93.71,93.62,93.03,89.88,92.64,92.53,n,92.09,92.80,88.83
        10\%,93.72,93.63,93.04,89.80,92.62,92.55,n,92.00,92.72,90.66
        20\%,93.77,93.66,93.02,89.93,92.63,92.48,n,92.02,92.84,88.64
        30\%,93.59,93.58,92.98,89.89,92.66,92.53,n,92.07,92.63,87.16
        40\%,93.09,93.45,93.03,89.75,92.70,92.63,n,92.12,92.80,85.31
        50\%,92.20,93.07,92.99,89.72,92.65,92.54,n,91.84,92.29,80.94
        60\%,90.46,90.81,92.66,89.84,92.70,92.55,n,91.48,91.20,69.48
        70\%,81.88,72.29,92.22,89.80,92.66,92.56,n,90.23,86.30,46.19
        80\%,32.02,19.84,84.03,85.80,91.86,92.54,n,83.44,63.00,23.71
        90\%,14.63,10.05,28.27,27.04,67.58,92.48,n,15.16,21.08,12.55
} \\
\vspace{0.2cm}
\begin{center}
\textbf{Unit Dropout/Pruning} \\
\end{center}
\hspace{3em}
\rotatebox[origin=c]{90}{\small prune percentage}\pgfplotstabletypeset[color cells resize65]{
        x,none,u25,t5050,t7533,t7566,t7590,n,var_unit,$L^1_{0.01}$,$L^0_{0.01}$
        0 \%,93.69,92.43,92.21,90.46,89.38,89.78,n,93.14,93.31,93.35
        10\%,90.05,67.52,91.96,88.44,89.48,90.18,n,92.91,91.03,83.01
        20\%,80.34,25.05,91.63,83.55,88.89,89.79,n,90.38,85.63,54.59
        30\%,59.94,13.47,91.30,69.82,88.84,89.88,n,86.38,72.19,21.34
        40\%,35.40,10.02,89.89,54.42,87.54,89.98,n,83.59,46.41,10.82
        50\%,12.63,9.97,88.41,28.88,84.86,90.05,n,65.79,26.72,15.04
        60\%,10.65,9.99,26.55,18.55,81.98,90.08,n,41.05,12.11,9.46
        70\%,11.70,10.01,17.41,17.84,75.47,90.03,n,19.36,11.81,10.02
        80\%,9.99,9.95,10.63,10.87,28.99,34.18,n,9.56,14.73,14.88
        90\%,9.85,9.98,9.30,10.29,9.97,10.04,n,10.41,10.22,9.98
}
\vspace{0.7em}
\caption{ResNet-32 model accuracies on CIFAR-10 at differing pruning percentages and under different regularisation schemes. The top table depicts results using the weight pruning strategy, while the bottom table depicts the results of unit pruning (see Sec.~\ref{sec:pruning})}
\label{tab:resnet}
\end{table*}

In order to analyze the effects of targeted dropout we construct a toy experiment with small dense networks to analyse properties of the network's dependence on its weights. The model we consider is a single hidden layer densely connected network with ten units and ReLU activations \citep{nair2010rectified}. We train two of these networks on CIFAR-10; the first unregularised, and the second with targeted dropout applied to the \(\gamma=75\%\) lowest-magnitude weights at a rate of \(\alpha=50\%\). The networks are both trained for 200 epochs at a learning rate of 0.001 using stochastic gradient descent without momentum. 

We then compute the gradient and Hessian over the test set in order to estimate the change in error from Equation~\ref{eqn:cie} (see Table~\ref{tab:alt}). In addition, we compute the Hessian-weight product matrix formed by typical element \([\theta^\top\odot\bm{H}\odot\theta]_{ij}=\theta_i\bm{H}_{ij}\theta_j\) as an estimate of weight correlations and network dependence (see Figure~\ref{fig:hess}). This matrix is an important visualisation tool since summing the entries associated with weights you intend to delete corresponds to computing the second term in Equation~\eqref{fig:hess} -- this becomes the dominant term towards the end of training, at which time the gradient is approximately zero.

Figure~\ref{fig:hess} makes clear the dramatic effects of targeted dropout regularisation on the network. In the Figure, we reorder the rows and columns of the matrices so that the first 25\% of the matrix rows/columns correspond to the 25\% of weights we identify as the important subnetwork (i.e. highest magnitude weights), and the latter 75\% are the weights in the unimportant subnetwork (i.e. lowest magnitude weights). The network trained with targeted dropout relies nearly exclusively on the 25\% of weights with the largest magnitude at the end of training. Whereas, the network trained without regularisation relies on a much larger portion of the weights and has numerous dependencies in the parameters marked for pruning.

\subsection{ResNet}\label{sec:resnet}

We test the performance of targeted dropout on Residual Networks (ResNets) \citep{he2016deep} applied to the CIFAR-10 dataset, to which we apply basic input augmentation in the form of random crops, random horizontal flipping, and standardisation. This architectural structure has become ubiquitous in computer vision, and is gaining popularity in the domains of language \citep{kalchbrenner2016neural}, and audio \citep{van2016wavenet}. Our baseline model reaches over 93\% final accuracy after 256 epochs, which matches previously reported results for ResNet-32 \citep{he2016deep}. 

Our weight pruning experiments demonstrate that standard dropout schemes are comparatively weak compared to their targeted counterparts; standard dropout performs worse than our no-regularisation baseline. We find that a higher targeted dropout rate applied to a larger portion of the weights results in the network matching unregularised performance with only 40\% of the parameters.

Variational dropout seems to improve things marginally over the unregularised baseline in both weight and unit pruning scenarios, but was still outperformed by targeted dropout.
\(L^0\) regularisation was fairly insensitive to its complexity term coefficient; we searched over a range of \(\beta\in[10^{-6}, 10^1]\) and found that values above $10^{-1}$ failed to converge, while values beneath $10^{-4}$ tended to show no signs of regularisation. Similarly to variational dropout, \(L^0\) regularisation does not prescribe a method for achieving a specific prune percentage in a network, and so, an extensive hyperparameter search becomes a requirement in order to find values that result in the desired sparsity. As a compromise, we search over the range mentioned above and select the setting most competitive with targeted dropout; next, we applied magnitude-based pruning to the estimates provided in Equation~13 of \citet{louizos2017learning}. Unfortunately, \(L^0\) regularisation seems to force the model away from conforming to our assumption of importance being described by parameter magnitude. 

\begin{table}[t]
\hspace{.13\columnwidth}
\begin{subfigure}[c]{.35\columnwidth}
\begin{center}
\stackanchor{\textbf{Weight}}{\textbf{Dropout/Pruning}}
\end{center}
\hspace{0em}
\rotatebox[origin=c]{90}{\small prune percentage}\pgfplotstabletypeset[color cells resize75]{
        x,none,t5050,$L^1_{10^{-5}}$
        0 \%,75.9,75.7,70.6
        10\%,75.9,75.7,70.4
        20\%,74.9,75.3,69.8
        30\%,71.9,74.4,65.7
        40\%,64.4,73.5,62.1
        50\%,45.0,68.8,53.4
        60\%,8.6 ,50.5,38.3
        70\%,0.7 ,14.8,17.6
        80\%,0.2 ,0.4 ,1.4 
        90\%,0.1 ,0.1 ,0.4 
}
\vspace{0.7em}
\end{subfigure} %
\hspace{.08\columnwidth}
\begin{subfigure}[c]{.35\columnwidth}
\begin{center}
\stackanchor{\textbf{Unit}}{\textbf{Dropout/Pruning}}
\end{center}
\hspace{0em}
\rotatebox[origin=c]{90}{\small prune percentage}\pgfplotstabletypeset[color cells resize65]{
        x,none,t5050,$L^1_{0.001}$
        0 \%,75.7,74.3,75.7
        10\%,34.5,67.2,66.6
        20\%,1.8 ,59.4,12.6
        30\%,0.4 ,33.0,0.4
        40\%,0.1 ,6.4 ,0.2
        50\%,0.1 ,0.6 ,0.1
        60\%,0.1 ,0.2 ,0.1
        70\%,0.1 ,0.1 ,0.1
        80\%,0.1 ,0.1 ,0.1
        90\%,0.1 ,0.1 ,0.1
}
\vspace{0.7em}
\end{subfigure}

\caption{ResNet-102 model accuracies on ImageNet. Accuracies are top-1, single crop on 224 by 224 pixel images.}
\label{tab:imagenet}
\end{table}

In Table~\ref{tab:imagenet} we present the results of pruning ResNet-102 trained on ImageNet. We observe similar behaviour to ResNet applied to CIFAR-10, although it's clear that the task utilises much more of the network's capacity, rendering it far more sensitive to pruning relative to CIFAR-10.

\subsection{Wide ResNet}

In order to ensure fair comparison against the \(L^0\) regularisation baseline, we adapt the authors own codebase\footnote{the original \(L^0\) PyTorch code can be found at: \url{github.com/AMLab-Amsterdam/L0_regularization}} to support targeted dropout, and compare the network's robustness to sparsification under the provided \(L^0\) implementation and targeted dropout. In Table~\ref{tab:wideresnet} we observe that \(L^0\) regularisation fails to truly sparsify the network, but has a strong regularising effect on the accuracy of the network (confirming the claims of \citeauthor{louizos2017learning}). This further verifies the observations made above, showing that \(L^0\) regularisation fails to sparsify the ResNet architecture.

\begin{table}[h]
\begin{center}
\textbf{Unit Dropout/Pruning} \\
\end{center}
\hspace{12em}
\rotatebox[origin=c]{90}{\small prune percentage}\pgfplotstabletypeset[color cells resize75]{
        x,none,t7533,$L^0_{10^{-6}}$
        0 \%,92.21,92.24,94.15
        10\%,89.76,92.09,88.05
        20\%,82.37,91.55,65.03
        30\%,52.20,90.09,13.34
        40\%,18.48,87.47,10.01
        50\%,10.53,82.09,10.00
        60\%,10.04,69.58,10.00
        70\%,10.00,44.05,10.00
        80\%,10.00,16.94,10.00
        90\%,10.00,10.43,10.00
}
\vspace{1em}
\caption{Wide ResNet \citep{zagoruyko2016wide} model classification accuracy on CIFAR-10 test set at differing prune percentages.}
\label{tab:wideresnet}
\end{table}

\subsection{Transformer}

\begin{table}[ht]
\begin{subfigure}[c]{.45\columnwidth}
\begin{center}
\textbf{Weight Dropout/Pruning} \\
\end{center}
\hspace{1em}
\rotatebox[origin=c]{90}{\small prune percentage}\pgfplotstabletypeset[color cells resize75]{
        x,none,t7566,t9066
        0 \%,26.01,26.52,25.32
        10\%,26.05,26.44,25.32
        20\%,25.90,26.48,25.19
        30\%,25.91,26.30,25.27
        40\%,25.81,26.20,24.97
        50\%,25.08,26.03,24.93
        60\%,23.31,25.62,24.27
        70\%,8.89,24.07,22.41
        80\%,0.24,12.39,10.57
        90\%,0.01,0.07,0.64
}
\caption{Transformer model uncased BLEU score.}
\label{tab:transformerbleu}
\end{subfigure} %
\hspace{.08\columnwidth}
\begin{subfigure}[c]{.45\columnwidth}
\begin{center}
\textbf{Weight Dropout/Pruning} \\
\end{center}
\hspace{1em}
\rotatebox[origin=c]{90}{\small prune percentage}\pgfplotstabletypeset[color cells resize75]{
        x,none,t7566,t9066
        0 \%,62.29,58.31,57.41
        10\%,62.54,59.00,58.10
        20\%,62.21,59.39,58.52
        30\%,62.33,58.66,57.86
        40\%,61.81,59.39,58.67
        50\%,60.82,57.71,57.08
        60\%,58.13,58.42,57.96
        70\%,48.40,55.39,54.85
        80\%,25.80,47.09,46.63
        90\%,6.90,21.64,27.02
}
\caption{Transformer model per-token accuracy.}
\label{tab:transformeracc}
\end{subfigure}
\caption{Evaluation of the Transformer Network under varying sparsity rates on the WMT newstest2014 EN-DE test set.}
\end{table}

The Transformer network architecture \citep{transformer} represents the state-of-the-art on a variety of NLP tasks. In order to evaluate the general applicability of our method we measure the Transformer's robustness to weight-level pruning without regularisation, and compare this against two settings of targeted dropout applied to the network.

The Transformer architecture consists of stacked multi-head attention layers and feed-forward (densely connected) layers, both of which we target for sparsification; within the multihead attention layers, each head of each input has a unique linear transformation applied to it, which are the weight matrices we target for sparsification.

Table~\ref{tab:transformerbleu} details the results of pruning the Transformer architecture applied to the WMT newstest2014 English-German (EN-DE). Free of any regularisation, the Transformer seems to be fairly robust to pruning, but with targeted dropout we are able to increase the BLEU score by 15 at 70\% sparsity, and 12 at 80\% sparsity; further confirming target dropout's applicability to a range of architectures and datasets.

\subsection{Scheduling the Targeting Proportion}
\label{rampingtd}

\begin{table}[h]
\begin{center}
\textbf{Weight Dropout/Pruning} \\
\end{center}
\hspace{3em}
\rotatebox[origin=c]{90}{\small prune percentage}\pgfplotstabletypeset[color cells resize75]{
        x,t7566,smallify1en51,n,m,smallify1en52,ramptarg1,o,m2,ramptarg2
        0 \%,92.64,90.16,n,90\%,90.20,89.03,n,98.5\%,89.03
        10\%,92.62,90.13,n,91\%,90.33,89.16,n,98.6\%,89.08
        20\%,92.63,90.16,n,92\%,90.30,89.14,n,98.7\%,89.00
        30\%,92.66,90.06,n,93\%,90.27,89.03,n,98.8\%,89.05
        40\%,92.70,90.17,n,94\%,89.46,89.05,n,98.9\%,88.99
        50\%,92.65,90.20,n,95\%,89.41,89.05,n,99.0\%,89.10
        60\%,92.70,90.12,n,96\%,88.55,89.02,n,99.1\%,88.35
        70\%,92.66,90.10,n,97\%,86.35,89.05,n,99.2\%,79.88
        80\%,91.86,90.15,n,98\%,59.27,89.05,n,99.3\%,77.35
        90\%,67.58,90.16,n,99\%,13.83,88.97,n,99.4\%,16.55
} \\
\vspace{0.2cm}
\begin{center}
\textbf{Unit Dropout/Pruning} \\
\end{center}
\hspace{11em}
\rotatebox[origin=c]{90}{\small prune percentage}\pgfplotstabletypeset[color cells resize75]{
        x,t7566,smallify1en4,ramptarg3
        0 \%,90.55,90.20,85.98
        10\%,90.83,90.33,86.12
        20\%,89.88,90.30,86.01
        30\%,87.35,90.27,86.10
        40\%,85.39,89.46,85.98
        50\%,80.84,89.41,86.13
        60\%,71.97,88.55,86.02
        70\%,55.98,86.35,86.08
        80\%,10.02,59.27,85.95
        90\%,10.07,13.83,85.99
}
\vspace{1em}
\caption{Comparing Smallify to targeted dropout and ramping targeted dropout. Experiments on CIFAR10 using ResNet32.}
\label{tab:smallify}
\end{table}

Upon evaluation of weight-level Smallify \citep{leclerc2018smallify} we found that, with tuning, we were able to out-perform targeted dropout at very high pruning percentages (see Table~\ref{tab:smallify}). One might expect that a sparsification scheme like Smallify -- which allows for differing prune rates between layers -- would be more flexible and better suited to finding optimal pruning masks; however, we show that a variant of targeted dropout we call \textit{ramping targeted dropout} is capable of similar high rate pruning. Moreover, ramping targeted dropout preserves the primary benefit of targeted dropout: fine control over sparsity rates.

Ramping targeted dropout simply anneals the targeting rate \(\gamma\) from zero, to the specified final \(\gamma\) throughout the course of training. For our ResNet experiments, we anneal from zero to 95\% of \(\gamma\) over the first forty-nine epochs, and then from 95\% of \(\gamma\) to 100\% of \(\gamma\) over the subsequent forty-nine. In a similar fashion, we ramp \(\alpha\) from 0\% to 100\% linearly over the first ninety-eight steps.

Using ramping targeted dropout we are able to achieve sparsity of 99\% in a ResNet32 with accuracy 87.03\% on the CIFAR-10 datatset; while the best Smallify run achieved intrinsic sparsity of 98.8\% at convergence with accuracy 88.13\%, when we perform pruning to enforce equal pruning rates in all weight matrices, the network degrades rapidly (see Table~\ref{tab:smallify}).

\subsection{Fixed Filter Sparsity}

We also propose a variation of Ramping targeted dropout (Section~\ref{rampingtd}), where each layer is assigned a \(\gamma\)\ such that only a fixed number of weights are non-zero by the end of training (for example, three parameter per filter). We refer to this as Xtreme dropout. ResNet32 when trained with Xtreme-3 (3 weights per filter are non-zero) was able to achieve an accuracy of 84.7\% on the CIFAR-10 datatset at a sparsity level of 99.6\% while Xtreme-4 was able to achieve 87.06\% accuracy at a sparsity level of 99.47\%. An interesting observation of Xtreme pruning is that when trained on ResNet18, it achieves 82\% accuracy at a sparsity level of 99.8\%. When translated to the number of parameters, it has only 29,760 non-zero parameters (includes BatchNorm) which is less than the number of parameters in a network consisting of a single dense layer with 10 output units.

\section{Exploring Recent Discussions and Concerns}

A line of work \citep{crowley2018pruning,liu2018rethinking} has suggested that post hoc pruning with fine-tuning is not as effective as it could be. They propose using the sparsity patterns derived from a pruned model to define a smaller network (where the remaining, unpruned weights are reinitialised randomly) which is then trained from scratch, yielding better final task performance than fine-tuning the pruned model's weights.

A similar question arose in our own work as we pondered how early in the training procedure the important subnetwork could be decided. In the ideal case, the important subnetwork would be arbitrary and we could blindly select any subnetwork at the beginning of training, delete the remaining network, and recover similar accuracy to a much more complicated pruning strategy. While in the worst case, the important subnetwork would be predestined, and would remain difficult to identify until the very end of training.

While \citet{crowley2018pruning,liu2018rethinking} rely on sparsity patterns derived from pruned models, in this paper we are concerned with pruning schemes that achieve sparsity in a single execution of the training procedure; and so, in order to evaluate the more general claim that training smaller networks from scratch can match (or even out-perform) pruning, we compare the following two methods:
\begin{itemize}
    \item \textbf{Random-pruning:} Before training, prune away a random subnetwork.
    \item \textbf{Targeted Dropout (Ramping TD):} Apply ramping targeted dropout throughout the course of training.
\end{itemize}

\begin{table*}[ht]
  \centering
  \begin{subfigure}[c]{.47\columnwidth}
  \centering
  \resizebox{0.98\columnwidth}{!}{
  \begin{tabular}{ccccc}
    \toprule
    & \multicolumn{4}{c}{\textbf{Prune \%}} \\
    \cmidrule{2-5}
    \textbf{Type} & 50\% & 75\% & 90\% & 99\% \\
    \cmidrule[0.2mm]{1-5}
    Random-prune & 92.58 & 92.32 & 90.66 & 80.86 \\
    Ramping TD           & \textbf{93.29} & \textbf{92.72} & \textbf{92.51} & \textbf{88.80} \\
    \cmidrule[0.3mm]{1-5}
  \end{tabular}
  }
  \caption{Comparison of weight-level pruning methods using ResNet-32 trained on CIFAR-10.}
  \end{subfigure}%
  \hspace{.04\columnwidth}
  \begin{subfigure}[c]{.47\columnwidth}
  \centering
  \resizebox{0.98\columnwidth}{!}{
  \begin{tabular}{cccccc}
    \toprule
    & \multicolumn{4}{c}{\textbf{Prune \%}} \\
    \cmidrule{2-5}
    \textbf{Type} & 75\% & 85\% & 90\% & 95\% \\
    \cmidrule[0.2mm]{1-5}
    Random-prune & 90.50 & 88.52 & 84.98 & 79.09 \\
    Ramping TD           & \textbf{90.84} & \textbf{88.59} & \textbf{86.45} & \textbf{80.65} \\
    \cmidrule[0.3mm]{1-5}
  \end{tabular}
  }
  \caption{Comparison of unit-level pruning methods using ResNet-32 trained on CIFAR-10.}
  \end{subfigure}%
  \vspace{1em}
  \begin{subfigure}[c]{1.\linewidth}
  \centering
  \begin{tabular}{cccccc}
    \toprule
    & \multicolumn{4}{c}{\textbf{Prune \%}} \\
    \cmidrule{2-5}
    \textbf{Type} & 75\% & 85\% & 90\% & 95\% \\
    \cmidrule[0.2mm]{1-5}
    Random-prune & 48.98 (0.62) & 45.58 (1.25) & 40.50 (2.03) & 31.44 (1.64)\\
    Ramping TD   & \textbf{52.64} (0.61) & \textbf{49.20} (0.10) & \textbf{45.03} (0.83) & 30.15 (1.72) \\
    \cmidrule[0.3mm]{1-5}
  \end{tabular}
  \caption{Comparison of unit-level pruning methods using VGG-16 trained on CIFAR-100. Results are the average of five independent training runs followed by one standard deviation reported in brackets.}
  \end{subfigure}
  \caption{Comparison between random pruning at the beginning of training and regularising with targeted dropout throughout the course of training, followed by post hoc pruning.}
  \label{tab:earlyprune}
\end{table*}

The results of our experiment are displayed in Table~\ref{tab:earlyprune}. It is clear that -- although \citet{crowley2018pruning,liu2018rethinking}'s results show that knowing a \emph{good} sparsity pattern in advance allows you to achieve competitive results with pruning -- simply training a smaller subnetwork chosen at random does not compete with a strong regularisation scheme used over the course of training. Similar observations that contradict the conclusions of \citet{crowley2018pruning,liu2018rethinking} have been made in both \citet{franklelotterticketatscale} and \citet{galestateofsparsity}.

\section{Conclusion}

We propose \textit{targeted dropout} as a simple and effective regularisation tool for training neural networks that are robust to {\it post hoc} pruning. Among the primary benefits of targeted dropout are the simplicity of implementation, intuitive hyperparameters, and fine-grained control over sparsity - both during training and inference. Targeted dropout performs well across a range of network architectures and tasks, demonstrating is broad applicability. Importantly, like~\cite{rippel2014learning}, we show how dropout can be used as a tool to encode prior structural assumptions into neural networks. This perspective opens the door for many interesting applications and extensions.

\section*{Acknowledgements}
We would like to thank Christos Louizos for his extensive review of our codebase and help with debugging our \(L^0\) implementation, his feedback was immensely valuable to this work. Our thanks goes to Nick Frosst, Jimmy Ba, and Mohammad Norouzi who provided valuable feedback and support throughout.
We would also like to thank Guillaume Leclerc for his assistance in verifying our implementation of Smallify. 

\newpage
\bibliographystyle{plainnat}
\bibliography{references}

\end{document}